# SESR: Single Image Super Resolution with Recursive Squeeze and Excitation Networks

Xi Cheng, Xiang Li, Jian Yang
School of Computer Science and Engineering
Nanjing University of Science and Technology
Nanjing, China
opteroncx@gmail.com
{xiang.li.implus, csjyang}@njust.edu.cn

Ying Tai
Youtu Lab
Tencent
Shanghai, China
yingtai@tencent.com

*Abstract*—Single image super resolution is a very important computer vision task, with a wide range of applications. In recent years, the depth of the super-resolution model has been constantly increasing, but with a small increase in performance, it has brought a huge amount of computation and memory consumption. In this work, in order to make the super resolution models more effective, we proposed a novel single image super resolution method via recursive squeeze and excitation networks (SESR). By introducing the squeeze and excitation module, our SESR can model the interdependencies and relationships between channels and that makes our model more efficiency. In addition, the recursive structure and progressive reconstruction method in our model minimized the layers and parameters and enabled SESR to simultaneously train multi-scale super resolution in a single model. After evaluating on four benchmark test sets, our model is proved to be above the state-of-the-art methods in terms of speed and accuracy.

## I. INTRODUCTION

Single image super resolution (SISR) is a hot topic in computer vision and has high practical value in many fields such as video, photography, games and medical imaging. The task of super resolution is to restore the low-resolution (LR) image to high-resolution (HR) images. When the upscaling factor is large, it is hard to learn the mapping from LR to HR and restore visual pleasing images. In recent years, neural networks were utilized in super resolution and showed great improvement in the reconstruction quality. To gain better restoration performance the super resolution models become deeper and deeper by stacking convolutional layers, many models with the depth exceeding 80 layers have appeared. Although increasing the depth of the model spatially can improve the performance of super resolution quality, it will bring a huge amount of computation and memory consumption. In order to improve the efficiency of super resolution networks, inspired by the SENet [1] we proposed a novel single image super resolution method with recursive squeeze and excitation network named SESR.

The squeeze and excitation (SE) module is used to model the interdependencies among channels and reweight the features. The information among channels would be selected and the efficiency of the model is greatly improved. We found that after adding the SE structure, the model could achieve very high reconstruction performance only with few residual blocks. Although the SE structure adds a small number of weighting layers, the number of layers and parameters in SESR is far fewer than that of other models when achieving similar level of super resolution performance. As shown in figure 1, we designed the model with a recursive structure in which the data continuously pass through the recursive unit. In addition, our model is end to end which means our model can input the low-resolution images directly. Different to DRRN [2] and other previous methods, our model do not need a bicubic input and we used a deconvolution layer as the upsample module in SESR which could decrease extra computation. Moreover, for large upscaling factors, our model used a progressively reconstruction method which means our model first reconstruct the lower scale image from the LR and share the information to the larger branch. This method also enables us to train multi scale super resolution in a single model. We summarize our contribution in the following points:

- We proposed a novel method for single image super resolution via squeeze and excitation module and recursive structure. Our model is proved to be over state-of-the-art methods in scale x4 benchmark not only in accuracy but also in speed.

- We found that adding the squeeze and excitation module can significantly improve the model performance, at

least 0.1dB gain in PSNR in each test dataset compared with models removed the SE module.

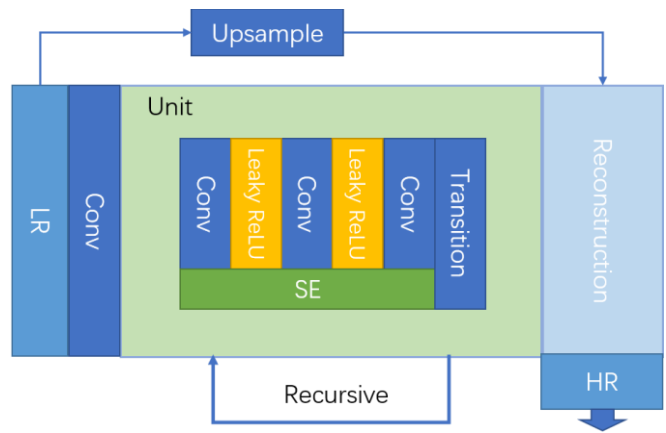

Fig. 1. Overview of our proposed model

- We designed the model with recursive structure and progressive reconstruction method which minimized the layers and parameters in the model.

## II. RELATED WORKS

### A. Classical super resolution methods

Super resolution is a hot topic in the field of computer vision. Although interpolation methods are widely used nowadays, the quality is hard to meet a satisfied level. Yang et al. [3-5] developed a set of super-resolution model based on sparse coding. Timofte et al. proposed A + [6] and IA [7] based on anchored neighborhood regression. Huang et al. proposed the SelfExSR [8] via transformed self-exemplars. The above methods achieved better results than bicubic but still hard to restore high quality images for higher upscaling factors.

### B. Deep learning based super resolution methods

In recent years, with the development of deep learning [9] and convolutional neural networks [10-11], many deep-learning-based super resolution methods have been proposed. The SRCNN [12] proposed by Dong et al. for the first time used convolutional neural networks on super resolution tasks. Simonyan et al. [13] found that the deepening of the network can bring about performance improvement. Then He et al. proposed ResNet [14] to make the deep models available for training. Inspired by the research above, the networks used in a super resolution tasks are also deepening. Kim et al. proposed a super resolution model with very deep convolution networks named VDSR [15] and later they proposed DRCN [16] with recursive structure. Tai et al. developed a 52-layer deep recursive network called DRRN [2] to further push super resolution performance. Lai et al. proposed LapSRN [17] by introducing the Laplacian pyramid which allows a network to do multiple-scale super-resolution simultaneously in one feed forward. To gain higher performance, more researches on the spatial structure were conducted. The networks are becoming more sophisticated instead of simply stacking the convolutional layers. Later MemNet [18] and SR DenseNet [19] which were designed to have different dense skip-connections [20] were proposed. The above methods are prone to be deeper and deeper. However, for some super resolution tasks the 84-layer MemNet [18] is not much better than the 52-layer DRRN [2], explosive growth of the size of the network could bring little improvement in super resolution quality but large amount of computation and GPU memory consumption.

### C. Perceptual loss and GANs

In order to make the image more visual pleasing, perceptual loss [21] was proposed and widely used in the style transfer [22] and super resolution field. In addition, Generative Adversarial Networks(GAN) can also produce visually beautiful images, recently many GAN based models were developed for single image super resolution such as SRGAN [23] and Neural Enhance [24]. Although GANs would produce good looking samples, the accuracy which evaluated by peak signal to noise ratio (PSNR) and structural similarity (SSIM) [25] is decreased compared with those supervised by L1 or L2 loss functions.

## III. PROPOSED METHODS

### A. Recursive structure

We build the model with a recursive structure which enables SESR to increase recursion depth without bring more parameters. In our model low resolution images are fed directly into the network and passed through a convolution layer, then entered the recursive unit. When output from squeeze and excitation enhanced residual blocks (SE-ResBlock), the features reinput to the recursive unit. Finally, the output entered the reconstruction network to obtain high resolution images. Different from DRCN [16], our model employed progressive reconstruction method and only get supervised at each end of different scale super resolution branch instead of supervised at each recursion which significantly decreased the amount of computation. Our model has local residual learning and global residual learning, the skip connections in our model makes it easy to converge. The total layers in one branch is 27 but we got higher performance than those deeper models.

### B. Squeeze and Excitation Residual Block

Inspired by the SENet [1], we introduced the SE module to make the super resolution models more efficient. Squeeze and excitation would make the network more powerful by emphasizing important features and suppress useless features among the channels. In order to squeeze global spatial information for each channel, we followed the SENet [1], using global average pooling [26] in our model. The squeeze function in SESR is shown as below:

$$x_c = S(u_c) = \frac{1}{HW} \sum_{i}^{H} \sum_{j}^{W} u_c(i,j) \tag{1}$$

Where $x_c$ is the c-th element of the squeezed channels and $S(.)$ is the squeeze function. $u_c$ is the c-th channel of the input. H and W denotes the height and width of the input.

The excitation function is shown as the following formula:

$$E(x) = \sigma(W_u(W_d x)) \tag{2}$$

Where $E(.)$ is the excitation function and x is the input squeezed signal from the previous layer. $\sigma$ denotes the Sigmoid, $W_u$ and $W_d$ denotes the 1x1 convolutional layer Conv Up and Conv Down shown in figure2.

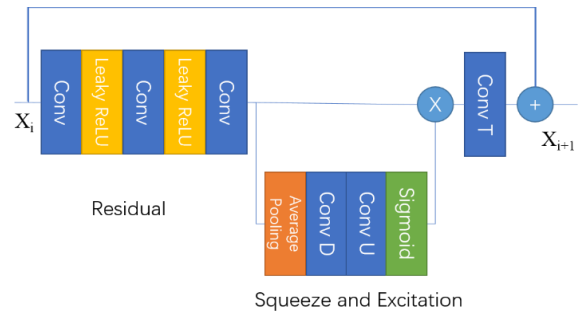

Fig. 2. We added squeeze and excitation into a common residual block

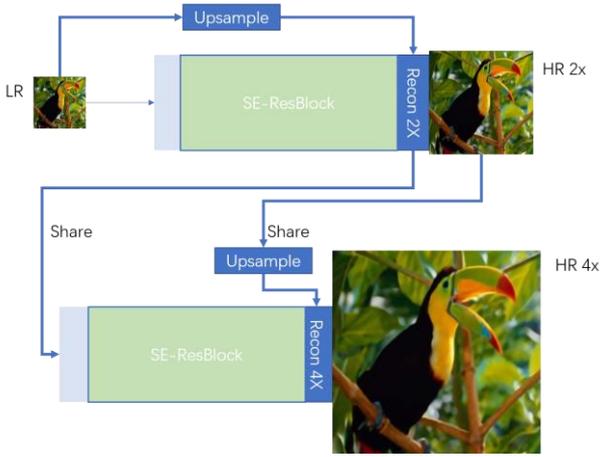

Fig. 3. The model could do multi scale upscaling task via Laplacian Pyramid

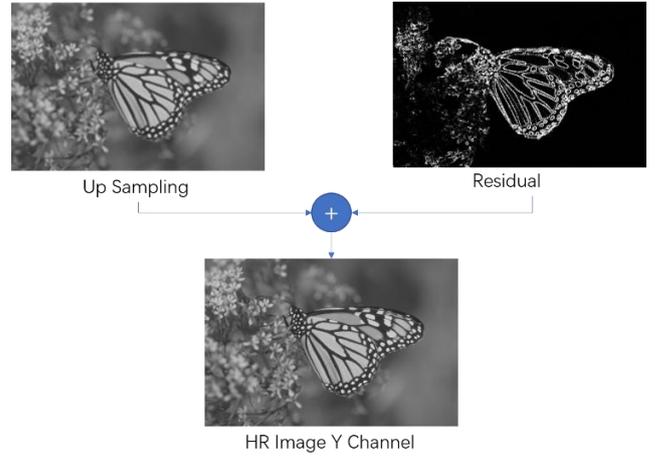

Fig. 4. Global residual learning for image reconstruction

We build our basic structure in the recursive unit on the base of residual blocks. Lim et al. [7] found that removing the batch normalization (BN) [27] layer would improve the performance of the super resolution network. We reproduced the experiment in the discussion section and we noticed the phenomenon. Therefore, we removed all the BN layers from the residual block which was proposed by He et al. [14]. and we added the squeeze and excitation module into the block. The SE-ResBlock we used in our SESR is shown in figure 2.

We first build the blocks by stacking convolutional layers interleaved with Leaky ReLU [28] then we put the SE module into the block. In contrast to original SENet [1], we used 1x1 convolutional layers instead of fully connected layers in the SE module. The number of channels in the first two convolution layers in each SE-Residual Block is 64, while the number of channels in the third convolution layer is increased by a factor of 4, followed by the SE module. In the SE module, squeeze was done by global average pooling. We used Conv Down to reduce the number of output channels to 16 and then Conv Up to increase the number of channels to 256 to form a Bottle Neck structure and then pass the sigmoid layer for modeling the correlations between the channels. The weights for channels were then multiplied with the residual. Finally, pass a 1x1 transition convolution to retransform the number of channels to 64, and add the output of the previous block to obtain $X_{i+1}$.

### C. Progressive Reconstruction

In order to improve the training efficiency of the model, a model needs to be trained with multiple scale for up sampling at the same time, inspired by the LapSRN [17], we introduced the laplacian pyramid in our model and the structure is shown as figure 3. LR images first input to the lower scale branch and reconstruct the HR 2x image then share the residual and image to the higher scale branch to reconstruct the HR 4x image.

Compared with those direct reconstruction methods, progressive methods could lead to better quality for higher upscaling factors and decreasing parameters by sharing information between each super resolution branch. Also, our progressive reconstruction enabled our model to do multi scale super resolution in a single model.

### D. Reconstruction Network

#### 1) Global residual learning

We first upscaling our low-resolution image via a deconvolution layer outside the recursive unit. The quality of the upscaled image from the tiny deconvolution layer is usually not very high but we used it to take the place of bicubic input. In the reconstruction network we add the low-quality upscaled image with the residual from the bottom recursive unit to obtain high-quality high-resolution image, the process is shown in Figure 4.

#### 2) Loss function

Even though directly optimizing the mean squared error (MSE) could get high peak signal to noise ratio (PSNR), the L2 loss always results in over smooth hard to restore visual pleasing images, so the Charbonnier Loss [17] is used as a loss function in the model. The loss function is shown as below:

$$\text{Loss} = \frac{1}{N}\sum_{i=1}^{N}\sum_{s=1}^{L}\sqrt{x^2 - \varepsilon^2}\,(\hat{y}_s^i - y_s^i) \qquad (3)$$

Where N is the batch size and set ε to 0.001, L is the number of up sampling branch. The s means the scale while the $y_s$ and $\hat{y}_s$ are the ground truth and generated high resolution image in a branch.

## IV. DISCUSSION

### A. Structure of residual blocks

As shown in figure 5, we researched on three different structures of residual blocks. Structure (a) is the same as the original ResNet [14] proposed by He et al. and this architecture was utilized in SR ResNet [23]. Structure (b) is the residual block similar to which was used in EDSR [29], all the BN layers were removed on the basis of structure (a), simplifying the network. Structure(c) is the SE-ResBlock we proposed in SESR which removed all the BN layer and added the SE module. The kernel size of convolutional layers in SE and the transition were set to 1x1 so it would not bring much parameters and that makes these three structures have similar number of parameters. Similarly, we put these three residual blocks into our recursive unit and set the recursion depth to four. Then we iterate 300 epochs on the same training set and test these trained models on Set5 [30] and Set14 [31]. The results are shown in Table 1.

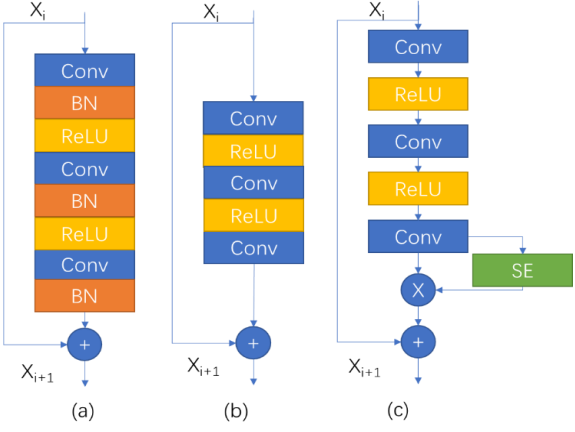

Fig.5. We researched on three different structure of residual blocks

As shown in the table, our SE-ResBlock achieves the highest score of PSNR and SSIM in both of the test datasets. When comparing PSNR, in Set5 our model is 0.22dB higher than (b) and 2.26 dB higher than (a). In Set15, our SESR is 0.1 dB higher than (b) and 1.08 dB higher than (a).

### B. Recursion depth

In this section, we study the effect of recursion depth with the model's reconstruction quality. We trained models with recursion depth of 2,3,4,5,6 with 391 training images and iterate for 300 epochs. We test these models for scale x4 on Set5 and Set14. The results are shown in Table 2.

Recursion depth directly impact the performance of the recursive network. As we can see in table 3, the reconstruction quality is highest when the recursion depth is set to four.

TABLE I. RECONSTRUCTION PERFORMANCE OF DIFFERENT STRUCTURES

| Model | PSNR/SSIM | |
|---|---|---|
| | Set5 | Set14 |
| Structure (a) | 29.58/ 0.840 | 27.24/ 0.753 |
| Structure (b) | 31.66/ 0.888 | 28.22/ 0.781 |
| Structure (c) | 31.84/ 0.891 | 28.32/ 0.784 |

TABLE II. NUMBER OF BLOCKS IN EACH SUPER RESOLUTION BRANCH

| Recursion Depth | PSNR/SSIM | |
|---|---|---|
| | Set5 | Set14 |
| 2 | 31.69/0.888 | 28.24/0.782 |
| 3 | 31.74/0.889 | 28.28/ 0.783 |
| 4 | 31.84/0.891 | 28.32/ 0.784 |
| 5 | 31.78/0.891 | 28.32/ 0.783 |
| 6 | 31.75/0.889 | 28.29/ 0.783 |

## V. EXPERIMENT

### A. Dataset

In this work we used Yang91 [4], BSD200 [32] and General100 [33] dataset for training. The model was evaluated with some public available and popular benchmark datasets including Set5 [30] and Set14 [31]. We also included the Berkeley segmentation dataset [32] (BSD100) and a dataset of urban landscape named Urban100 [8]. All the RGB images of these four benchmark datasets were converted to YCbCr color space with OpenCV, and we only input the Y channel to the network.

### B. Experiment Setup

In the experiment, we used a NVIDIA Tesla P40 for training our proposed models. We build the model using Pytorch version 0.2.0. The operating system of our server is Ubuntu16.10, CUDA8 and CUDNN5.1 were installed.

### C. Comparisons with state-of-the-art models

*1) Visual comparison*

Figure6 shows the reconstruction results and the ground truth from our test sets and we compare our proposed SESR with other state-of-the-art super resolution methods including A+ [6], SelfExSR [8], SRCNN [12], VDSR [15] and DRCN [16]. We cropped a 64x64 sub image from each reconstructed high-resolution images and compute the PSNR and SSIM of each sub image with the ground truth.

*2) Reconstruction Accuracy*

We use PSNR and SSIM as evaluation methods to evaluate the model on the above benchmark dataset. Same amounts of pixels of the border were ignored. The test images were first down sampling by bicubic and restored by the super resolution models.

The reconstruction quality for scale x2 and scale x4 of our SESR and other state-of-the-art models can be obtained from Table 3 and Table 4, we marked the best quality in red, the second in blue.

*3) Speed*

In this part, we researched on the running time of models. We reproduced LapSRN [17] and VDSR [15] with PyTorch. We test these methods on a Tesla P40 GPU. We tested on BSD100 [32] for scale x4. As shown in figure 9 SESR could run at a very high speed, less than 0.02 second per image and achieves the best accuracy among the state-of-the-art models.

*4) Model Parameters*

Both the SESR and LapSRN [17] contains two branches for different scale super resolution due to the progressive reconstruction method in the model. Our recursive model only contains 624k parameters. SESR is set with the recursion depth of 4. We compared the parameters and Set14 [31] results of SESR with other state-of-the-art models. From figure10, our proposed SESR is shown to be the most powerful model with a small number of parameters.

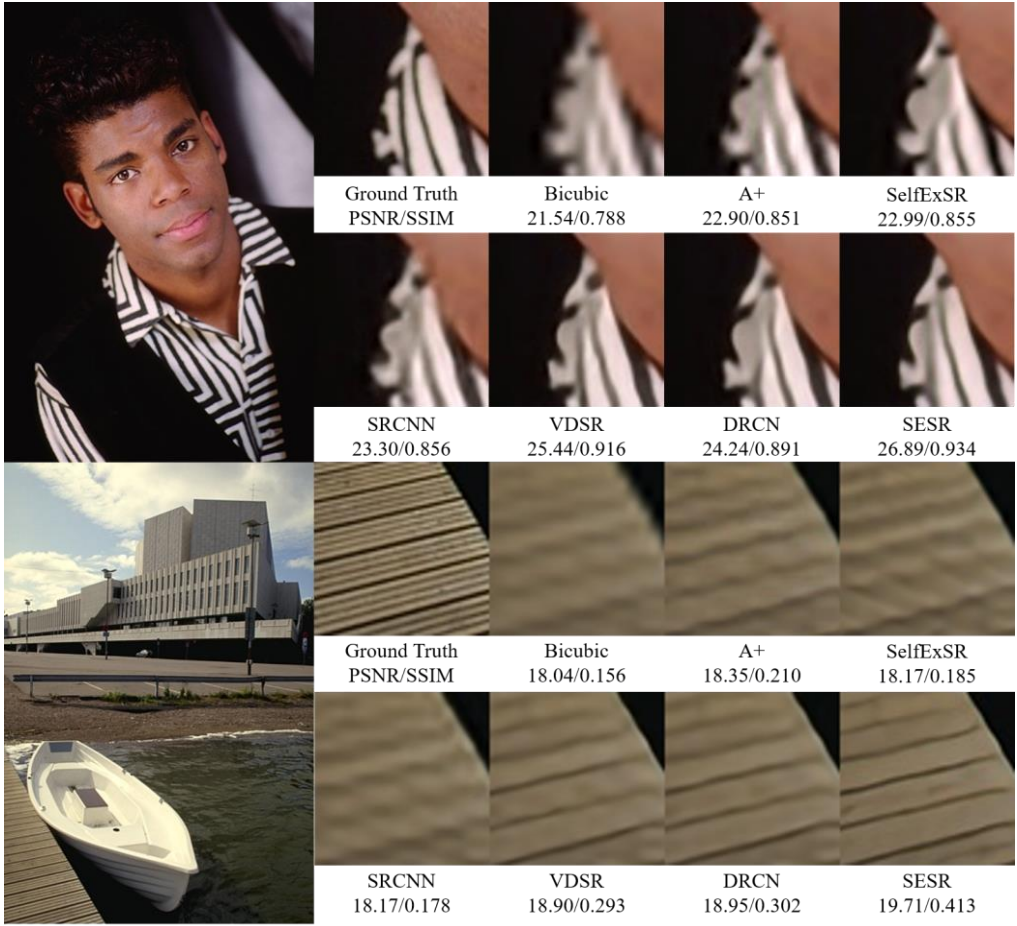

Fig.6. Quality comparison of our model with other work for scale x4 super resolution

TABLE III. PSNR AND SSIM FOR SCALE X4 ON SET5, SET14, BSD100 AND URBAN100

| Method | *PSNR/SSIM* | | | |
|---|---|---|---|---|
| | *Set5* | *Set14* | *Bsd100* | *Urban100* |
| Bicubic | 28.43/0.811 | 26.01/ 0.704 | 25.97/ 0.670 | 23.15/ 0.660 |
| A+ [6] | 30.32/ 0.860 | 27.34/ 0.751 | 26.83/ 0.711 | 24.34/ 0.721 |
| SRCNN [12] | 30.50/ 0.863 | 27.52/ 0.753 | 26.91/ 0.712 | 24.53/ 0.725 |
| FSRCNN [33] | 30.72/ 0.866 | 27.61/ 0.755 | 26.98/ 0.715 | 24.62/ 0.728 |
| SelfExSR [8] | 30.34/ 0.862 | 27.41/ 0.753 | 26.84/ 0.713 | 24.83/ 0.740 |
| VDSR [15] | 31.35/ 0.883 | 28.02/ 0.768 | 27.29/ 0.726 | 25.18/ 0.754 |
| DRCN [16] | 31.54/ 0.884 | 28.03/ 0.768 | 27.24/ 0.725 | 25.14/ 0.752 |
| LapSRN [17] | 31.54/ 0.885 | 28.19/ 0.772 | 27.32/ 0.727 | 25.21/ 0.756 |
| DRRN [2] | 31.68/ 0.888 | 28.21/ 0.772 | 27.38/ 0.728 | 25.44/ 0.764 |
| SESR(ours) | 31.84/ 0.891 | 28.32/ 0.784 | 27.42/ 0.737 | 25.42/ 0.771 |

## VI. CONCLUSION

In this study, we proposed a novel single image super resolution method. Compared with other methods, our model could achieve good results with fewer residual blocks and shallow recursion depth, effectively reducing the number of model parameters and calculating time. In addition, we also absorbed many excellent super resolution methods in the early stage and utilizing the progressive reconstruction methods so that our model could train higher scale better and could do a variety of super resolution scales in a single model. Our model was evaluated on serval testing datasets and we achieved the performance over the state-of-the-art methods not only in accuracy but also in speed.

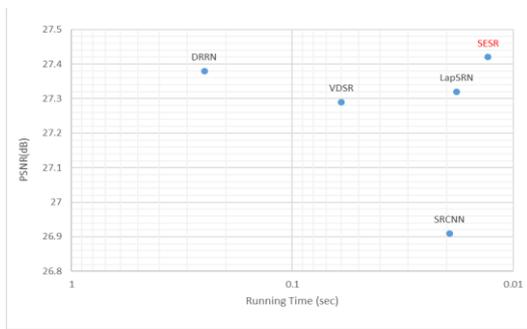

Fig. 7. Running time comparison with other state-of-the-art models

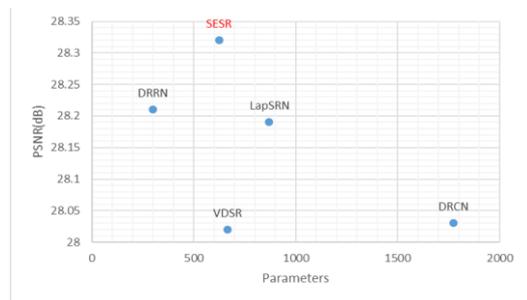

Fig. 8. Model parameters comparison with state-of-the-art models